\documentclass[letterpaper]{article} % DO NOT CHANGE THIS
\usepackage{aaai23}  % DO NOT CHANGE THIS
\usepackage{times}  % DO NOT CHANGE THIS
\usepackage{helvet}  % DO NOT CHANGE THIS
\usepackage{courier}  % DO NOT CHANGE THIS
\usepackage[hyphens]{url}  % DO NOT CHANGE THIS
\usepackage{graphicx} % DO NOT CHANGE THIS
\usepackage{natbib}  % DO NOT CHANGE THIS AND DO NOT ADD ANY OPTIONS TO IT
\usepackage{caption} % DO NOT CHANGE THIS AND DO NOT ADD ANY OPTIONS TO IT
\usepackage{algorithm}
\usepackage{algorithmic}

\usepackage{newfloat}
\usepackage{listings}
\usepackage{amsmath}
\usepackage{booktabs}
\DeclareCaptionStyle{ruled}{labelfont=normalfont,labelsep=colon,strut=off} % DO NOT CHANGE THIS
\floatstyle{ruled}
\newfloat{listing}{tb}{lst}{}
\floatname{listing}{Listing}
\title{Heterogeneous-Branch Collaborative Learning for Dialogue Generation}
\author {
    Yiwei Li,
    Shaoxiong Feng,
    Bin Sun,
    Kan Li\thanks{Corresponding author.}
}
\affiliations {
    School of Computer Science, Beijing Institute of Technology \\
    \{liyiwei,shaoxiongfeng,binsun,likan\}@bit.edu.cn
}

\usepackage{bibentry}
\begin{document}

\maketitle

\begin{abstract}
With the development of deep learning, advanced dialogue generation methods usually require a greater amount of computational resources. One promising approach to obtaining a high-performance and lightweight model is knowledge distillation, which relies heavily on the pre-trained powerful teacher.
Collaborative learning, also known as online knowledge distillation, is an effective way to conduct one-stage group distillation in the absence of a well-trained large teacher model.
However, previous work has a severe branch homogeneity problem due to the same training objective and the independent identical training sets. 
To alleviate this problem, we consider the dialogue attributes in the training of network branches.
Each branch learns the attribute-related features based on the selected subset.
Furthermore, we propose a dual group-based knowledge distillation method, consisting of positive distillation and negative distillation, to further diversify the features of different branches in a steadily and interpretable way.
The proposed approach significantly improves branch heterogeneity and outperforms state-of-the-art collaborative learning methods on two widely used open-domain dialogue datasets.
\end{abstract}

\section{Introduction}
Open-domain Neural dialogue generation \citep{Seq2Seq-Sordoni-2015,NoisyData-Vinyals-2015,Seq2Seq-ShangLifeng-2015}, aiming to generate diverse and coherent responses, has gained increasing attention and achieved impressive performance. 
It is important to recognize, however, that these considerable improvements typically come at the expense of over-parameterized networks, inhibiting their development on real-world resource-limited scenarios such as mobile chatbot applications.
Knowledge distillation is an appropriate knowledge-transfer methodology to resolve this issue, which uses predicted distributions \citep{DBLP:journals/corr/HintonVD15}, hidden states \citep{DBLP:conf/emnlp/SunCGL19}, or attention matrices \citep{DBLP:conf/emnlp/JiaoYSJCL0L20}, etc. of a teacher model as targets to induce the student to imitate.
A conventional distillation process involves two stages that begin with a cumbersome pre-trained teacher model and then distill the knowledge to the compact student model.
Unfortunately, training such a complex teacher model is time-consuming and a high-capacity model may not always be available.

\begin{figure}[th]
\centering
\includegraphics[width=1.0\linewidth]{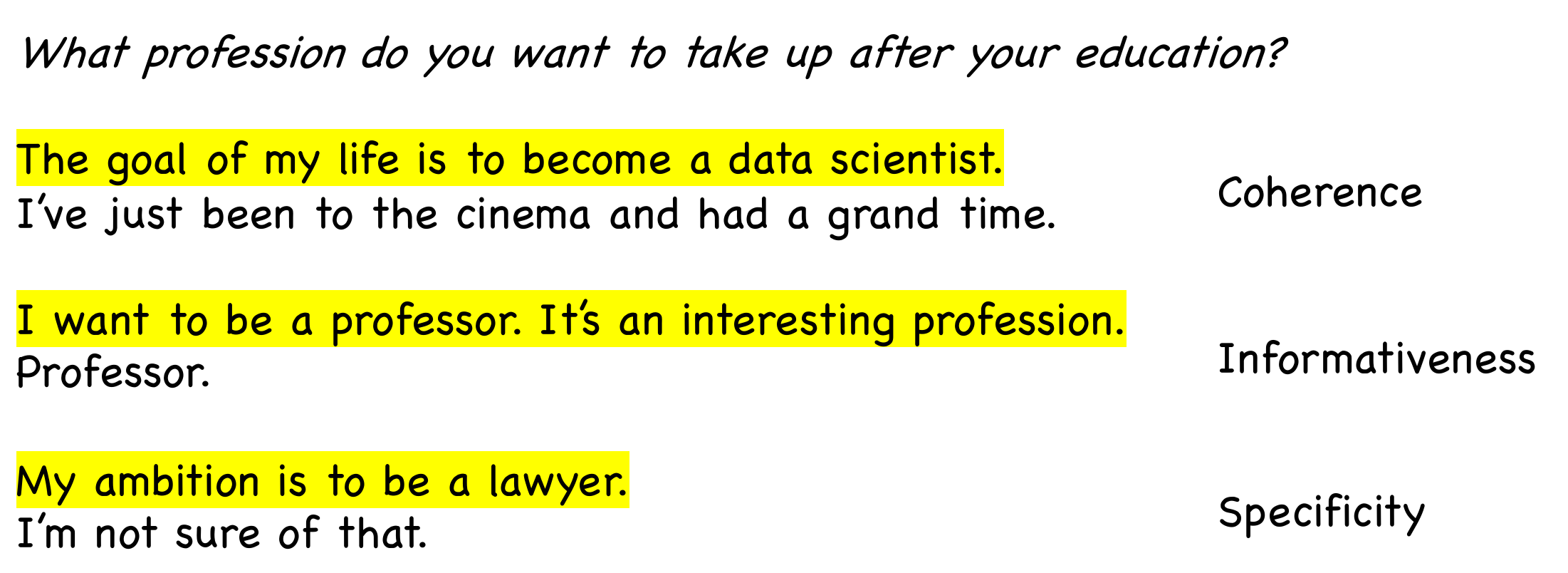}
\caption{Dialogue examples with two levels related to three dialogue attributes. The response quality can be assessed by multiple perspectives.}
\label{fig.example}
\end{figure} 

With a view to overcoming traditional limitations, online knowledge distillation \citep{Online-distill,ONE}, also called collaborative learning \citep{DML,CL-ILR}, is currently receiving considerable attention. 
Instead of pre-training a high-capacity teacher, collaborative learning conducts a single-stage group-based knowledge distillation that transfers the knowledge between less-parameterized student branches simultaneously. 
Aside from accelerating model learning efficiency over conventional KD, another major advantage of collaborative learning is the ability to find a more robust local minimum when compared to a single model learning method. 
It is important to note, however, that the homogeneity problem among branches along with training will lead to early saturation, thereby affecting the effectiveness of group distillation.
To alleviate this problem, \citet{OKDDip} impeded homogenization by equipping a diversity holding mechanism; \citet{PCL} randomly enhanced the input to guarantee the discrepancy between branches; \citet{CGL} proposed random routing to improve the diversity of features.

Even though the aforementioned approaches have demonstrated their superiority, one major drawback remains to limit further branch heterogenization: previous work only focused on the classification task, resulting in the same training objective (i.e. classification accuracy) for all branches of the framework with independent identical distribution (i.i.d.) training data. It will make different branches tend to converge to similar feature representations \citep{Convergent,ONE,OKDDip}. Consequently, there is barely any intuitive approach to allow branches to develop in various training directions, which is a significant obstacle to fostering diversity among them.

As opposed to classification task, goals of dialogue generation model focus more on dialogue attributes than the accuracy with the references. As demonstrated in \citet{FilterSpecificity-See-2019}, inadequate modeling of conversational aspects such as coherence and specificity results in inferior model performance and low response quality. Taking Figure~\ref{fig.example} as an example, response quality can be affected by multiple perspectives. As a result, directly applying the collaborative learning to the dialogue generation task will lead to sub-optimal performance. A variety of dialogue attributes can provide a natural insight into improving branch heterogeneity. With this in mind, it is possible to develop more effective and interpretable techniques to further enhance branch diversity.

In this work, we propose a heterogeneous attribute-aware collaborative learning paradigm for response generation, comprising two types of branches: auxiliary and master.
To achieve the goals of the dialogue system in a fine-grained way, we train each auxiliary branch on the corresponding aspect-specific sub-set to capture features along with some dialogue attributes. Each sub-set is collected according to the corresponding scoring method. Unlike auxiliary branches, the master branch is trained with the entire dataset to learn features roughly but comprehensively. 
In previous work, each branch learns from all the other branches, which is prone to homogenize different branches as the training continues. To further improve the diversity of auxiliary branches and integrate multi-view knowledge steadily, we propose the dual group-based knowledge distillation, consisting of positive distillation \citep{DBLP:journals/corr/HintonVD15} and negative distillation \citep{negative-distill}. 

Specifically, positive distillation is conducted from all the auxiliary branches to the master branch, transferring the attribute-specific knowledge effectively, whilst negative distillation is performed within the attribute-related branches to enforce them to learn different dialogue properties. 
Furthermore, negative distillation is implemented on hierarchical feature representation to use multi-level negative knowledge.
However, due to some common features shared by different auxiliary branches, blindly maximizing the distance of hidden states among auxiliary branches will harm the model performance and the training stability. To this end, we design a novel distillation approach called orthogonal negative distillation. It only strengthens the features of each auxiliary branch orthogonal to the other branches, avoiding disturbing the learning of common knowledge.

In summary, our contributions are as follows:
\begin{itemize}
    \item To the best of our knowledge, we are the first to propose the collaborative dialogue learning approach that transfers attribute-aware knowledge in a one-stage manner.
    \item Dual group-based knowledge distillation is proposed for better guiding auxiliary branches to learn attribute-specific knowledge. Orthogonal negative distillation can incentivize branches to capture biased features while avoiding harming the common knowledge.
    \item Extensive Evaluations on two widely used open-domain dialogue datasets demonstrate that the proposed approach significantly improves the branch heterogeneity and outperforms the state-of-the-art collaborative learning methods.
\end{itemize}
\section{Method}
\subsection{Approach Overview}
\begin{figure*}[th]
\centering
\includegraphics[width=0.95\linewidth]{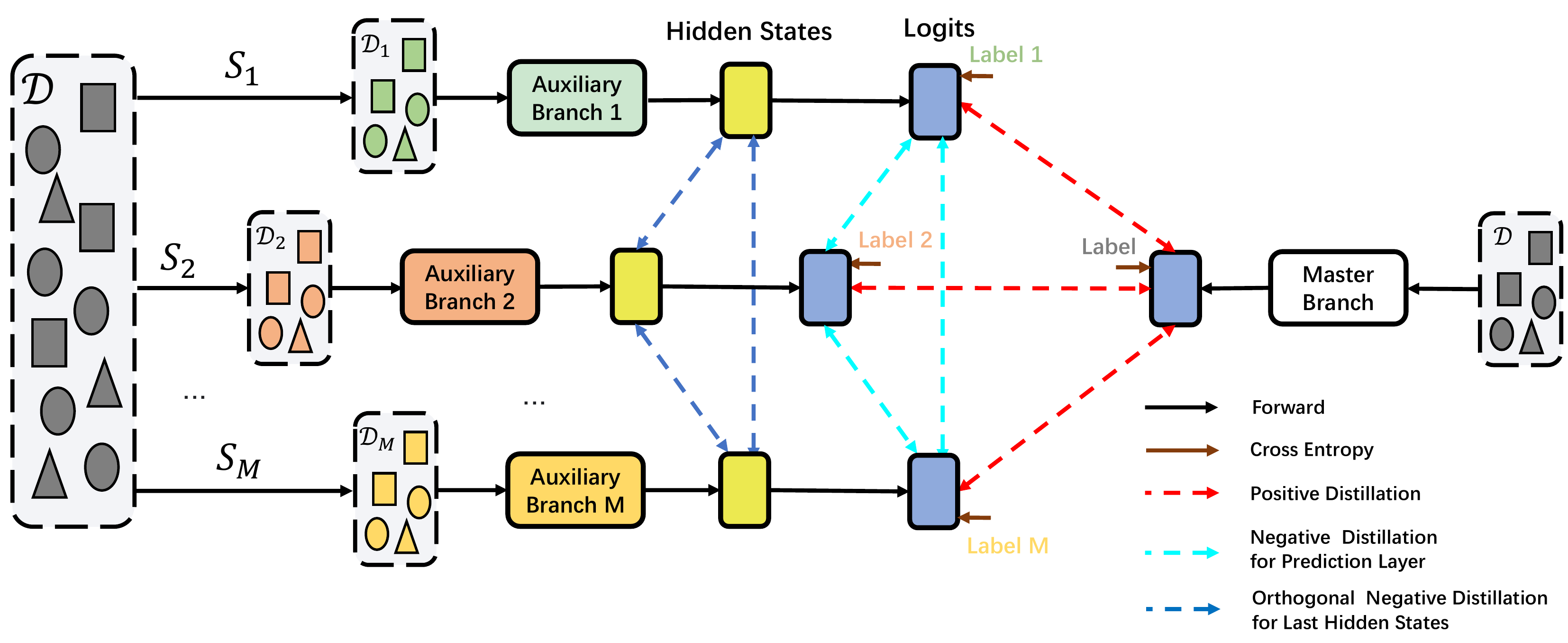}
\caption{An overview of the proposed heterogeneous attribute-aware collaborative dialogue learning.}
\label{fig.method1}
\end{figure*}

Taking $C=\{c_1, c_2,...,c_{T_{c}}\}$ as context, the objective of dialogue generation task is to generate the response $R=\{r_1, r_2,...,r_{T_{r}}\}$, where $T_{c}$ and $T_{r}$ represent the length of context and response, respectively. Instead of training a complicated and huge model, we build a collaborative dialogue learning framework to obtain a less-parameterized but effective model for inference. The overview of the proposed framework is illustrated in Figure~\ref{fig.method1}. 
In consideration of diverse dialogue attributes, we split the training dataset to several sub-sets according to scoring methods measuring the sample quality from multiple perspectives. Each attribute-related sub-set guides one branch to learn the corresponding specific knowledge. After that, we propose dual knowledge distillation in which positive distilling occurs between the master branch and all of the auxiliary branches, while negative distillation occurs within the attribute-related branches to encourage them to learn different dialogue properties. The orthogonal negative distillation is designed to identify biased features without interfering with knowledge.

\subsection{Dialogue Attribute Learning}

The generative dialogue model aims to learn a conditional probability distribution $p_{\theta}(R|C)$.
The maximum likelihood estimation (MLE) is usually used to guide the model to generate the target responses:
\begin{equation}
\mathcal{L}_{\mathrm{MLE}}=-\sum_{i=1}^{T_{r}} \log p_{\theta}\left(r_{i} \mid r_{<i}, C \right),
\label{eq:mle}
\end{equation}
where $r_{i}$ is the ground-truth tokens.
Therefore, the performance of the dialogue model largely depends on the distribution characteristics of the training set. 
Recently, a line of work introduces a data manipulation strategy, to boost the model performance with the corresponding dialogue attributes. 
They first measure the quality of samples in terms of a certain dialogue attribute by a scoring method, and then discard the low-score samples. The selection data can induce the model to learn attribute-related features more effectively for the generation of high-quality responses.
Specifically, the raw training samples $\mathcal{D}$ are reorganized into multiple view-specific training sub-sets ($\mathcal{D}_{1}, \mathcal{D}_{2}, \cdots, \mathcal{D}_{M}$) based on the scores of $\mathcal{S}_{m}$ and a certain selection proportion. Note that each sample can be assigned to multiple sub-sets as it may obtain high scores from more than one scoring method. 
Then, each branch $m$ is trained with corresponding sub-set $\mathcal{D}_{m}$ with Equation~\ref{eq:mle}.
Three dialogue attributes are considered in this paper and the following is the details of their corresponding scoring methods:
\paragraph{Coherence} reflects how well a dialogue response semantically relates to its context.
A joint score \citep{FilterConsistency-Akama-2020}:
\begin{equation}
S_{C+R}(c, r) = \alpha S_{C} + \beta S_{R}
\end{equation}
that contains two parts: connectivity $S_{C}$ and content relatedness $S_R$.
    The $S_{C}$ is evaluated by the co-occurrence of key-phrases ($p\in q$, $h\in r$):
    \begin{equation}
    % \small
    \label{eq:sc}
         S_C = 
         \sum_{(p,h)} \frac{\max(nPMI(p,h),0)\cdot |p| \cdot |h|} {|c| \cdot |r|},
    \end{equation}
    where $|\cdot|$ means the number of words and the $nPMI$ represents the normalized pointwise mutual information \citep{nPMI-Bouma-2009}. In addition, $S_{R}$ is evaluated by the cosine of the context and its response:
    \begin{equation}
    % \small
        S_R = \max(cos(c_{emb}, r_{emb}),0)
    \end{equation}
    The $c_{emb}$ and the $r_{emb}$ are vector representations of the context and response.
    % This scoring method can reflect the consistency of a dialogue pair. 
    
\paragraph{Informativeness} reflects how much the information related to the query is contained in the generated response, which is evaluated by Entropy\_Src \citep{FilterEntropy-Csaky-2019}: 
This score is the entropy of a response utterance:
    \begin{equation}
    % \small
        H_{src}(r|D) = -\sum_{(c_i, r)\in D}p(c_i|r)\log{p(c_i|r)},
    \end{equation}
    where $r$ represents the response, $D$ represents the dialogue dataset, and $c_i$ means a context of $r$ in $D$. By using this scoring method, the dialogue pair with many-to-one problem will be filtered, thereby alleviating the phenomenon of general response.
    
\paragraph{Specificity} \citep{FilterSpecificity-See-2019} reflects how much the generated response is good at word usage:
    \begin{equation}
    % \small
        Spe(t) = \frac{idf(t)-min\_idf}{max\_idf-min\_idf},
    \end{equation}
    where $t$ is a token of the response, and $idf(t)=\log(\frac{R}{R_t})$. $R$ is the number of responses in the dataset, and $R_t$ is the number of those responses that contain $t$. Using this scoring method, specific tokens can be identified in the response.

\subsection{Dual Knowledge Distillation}
The distillation objective is employed to alter the representation of two models, denoted as $f^{A}(x)$ and $f^{B}(x)$:

\begin{equation}
\mathcal{L}_{\mathrm{KD}}=\sum_{x \in \mathcal{D}} L\left(f^{A}(x), f^{B}(x)\right),
\end{equation}
where $L(\cdot)$ provides a measurement function for calculating distances between representations in  multi-levels.

A conventional collaborative learning process only distills positive knowledge, where $L(\cdot)$ is aiming to minimize the distance between branches. However, when it comes to attribute-related branches in dialogue learning, there are different directions in which they tend to converge. It is not appropriate to directly apply positive knowledge distillation to the collaborative dialogue learning framework. In this paper, we propose dual knowledge distillation consisting both positive and negative distillation (where $L(\cdot)$ seeks to maximize the distance between auxiliary branches), as a means of transferring attribute-specific knowledge in a reasonable manner.

\paragraph{Positive Distillation}
In order to transfer the attribute-aware knowledge to master branch, positive distillation (PD) is performed on the prediction layer:
\begin{align}
\nonumber  \mathcal{L}_{PD}(\boldsymbol{A}, \boldsymbol{B})  & =  -\sum_{i=1}^{T_{r}} \sum_{k=1}^{|\mathcal{V}|} p_{\boldsymbol{A}} \left(r_{i}=k \mid r_{<i}, C\right)  \\
 & \cdot \log p_{\boldsymbol{B}}\left(r_{i}=k \mid r_{<i}, C \right),
\end{align}
where $\boldsymbol{A}$, $\boldsymbol{B}$ refers to two branches and $p_{\boldsymbol{A}}$, $p_{\boldsymbol{B}}$ are calculated by:
\begin{equation}
p_{i}=\frac{\exp \left(z_{i} / T\right)}{\sum_{j} \exp \left(z_{j} / T\right)},
\end{equation}
where the probability distribution over words is softened with a temperature coefficient $T$.
Positive Distillation is carried out in a bidirectional manner between the master branch and the auxiliary branches. On the one hand, the attribute-specific knowledge can be absorbed by the master branch. On the other hand, the consolidated knowledge from the master branch needs to be transferred to the auxiliary branches in order to facilitate the generation of higher quality responses from them.

\paragraph{Negative Distillation for Prediction Layer}
For the purpose of encouraging auxiliary branches to better obtain its own specific knowledge, we use the soft unlikelihood loss from \citet{negative-distill} to achieve the negative distillation (ND) within them for the prediction layer first:
\begin{align}
\label{eq:pred}
\nonumber   \mathcal{L}_{ND_{pred}}(\boldsymbol{A}, \boldsymbol{B}) = & -\sum_{i=1}^{T_{r}} \sum_{k=1}^{|\mathcal{V}|} p_{\boldsymbol{B}} \left(r_{i}=k \mid r_{<i}, C \right) \\
& \cdot \log \left(1-p_{\boldsymbol{A}}\left(r_{i}=k \mid r_{<i}, C \right)\right),
\end{align}
Through this function, the distance between token prediction probabilities becomes larger, resulting in different branches producing different responses reflecting their own dialogue attributes and improving branch heterogeneity.

\paragraph{Negative Distillation with Orthogonal Projection}
Besides the explicit knowledge from the prediction layer, implicit knowledge embedded in the hidden states can also help the negative distillation process. In spite of the fact that different auxiliary branches acquire different attribute-specific knowledge, there should be some shared features in hidden states to support the basic abilities of sentence generation. Directly increasing the distance of hidden states between branches by negative distillation will damage the common knowledge for dialogue generation.

\begin{figure}[t]
\centering
\includegraphics[width=0.6\linewidth]{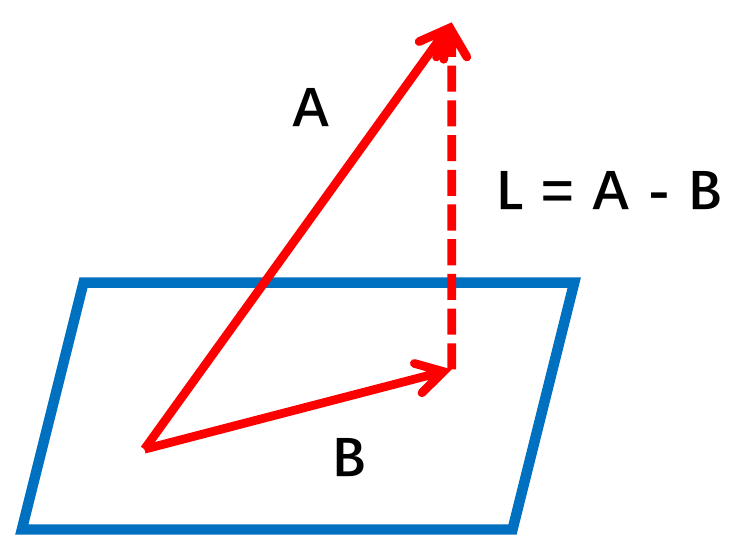}
\caption{Orthogonal Projection for Hidden States.}
\label{fig.method2}
\end{figure} 

Therefore, we propose orthogonal negative distillation to protect the common features from interference inspired by \citet{DBLP:conf/iclr/WangHLX19}. 
Specifically, as shown in Figure~\ref{fig.method2}, we project the hidden state $\mathbf{H}_{A}$ to the orthogonal space of hidden state $\mathbf{H}_{B}$ in order to get $\mathbf{H}_{L}$:
\begin{equation}
\mathbf{H}_{L}=\left(\mathbf{I}-\mathbf{H}_{B}\left(\mathbf{H}_{B}^{T} \mathbf{H}_{B}\right)^{-1} \mathbf{H}_{B}^{T}\right) \mathbf{H}_{A}
\end{equation}
$\mathbf{H}_{L}$ contains the biased features of $\mathbf{H}_{A}$ which reflects its attribute-specific knowledge comparing with $\mathbf{H}_{B}$, getting rid of the shared features within them.
On this basis, we conduct negative distillation with mean reverse square error (MRSE) \citep{negative-distill} between $\mathbf{H}_{L}$ and $\mathbf{H}_{B}$, which is conducive to dialogue attribute learning for branch $\boldsymbol{A}$ while avoiding the common knowledge interface. The loss function is then defined as:
\begin{equation}
    \mathcal{L}_{ND_{hidden}}(\mathbf{H}_{L}, \mathbf{H}_{B}) = \frac{1}{n} \sum^{n}_{i=1} \exp^{-SE(\mathbf{H}_{L}, \mathbf{H}_{B})},
\end{equation}
where $SE$ refers to square error. Note that we only perform ND on the last hidden states of decoder for training efficiency.

\paragraph{Optimization}
For the proposed collaborative dialogue learning framework, the overall objective function consists of two terms: a conventional cross entropy loss for dialogue generation and online knowledge distillation loss for collaborative learning.
Specifically, the loss for master branch is:
\begin{equation}
\mathcal{L}= \mathcal{L}_{MLE} + \frac{1}{|M|} \sum^m \mathcal{L}^m_{PD},
\end{equation}
where $|M|$ is the number of auxiliary branches.
While for each auxiliary branch:

\begin{align}
\nonumber   \mathcal{L}= \mathcal{L}_{MLE} +  \mathcal{L}^m_{PD}  + \frac{1}{|M|-1} \sum^{m} \mathcal{L}^m_{ND_{pred}} \\
 + \frac{1}{|M|-1} \sum^{m} \mathcal{L}^m_{ND_{hidden}}.
\end{align}
All branches are trained simultaneously at each epoch until the master branch converges. 
\section{Experiment}
\subsection{Datasets}
We evaluate the proposed method using two widely used dialogue datasets: 
\textbf{DailyDialog}, a collection of conversations that represent human daily communication \citep{dailydialog2017}, and \textbf{OpenSubtitles}, which consists of large-scale dialogues extracted from movie subtitles \citep{opensubtitles2009}. After data preprocessing, the number of context-response pairs in training/validation/test set is 68,066/6,820/6,841 for DailyDialog, and 200,000/20,000/10,000 for OpenSubtitles.

\begin{table*}[th]
    \centering
    \begin{tabular}{l c c c c c c c c c c c c}
    \toprule
    Models & Dist-1 & Dist-2 & Dist-3 & BLEU-1 & BLEU-4 & AVE & COH & H-1 & H-2 & H-3 & KL & LF 
    \\ \midrule
       Transformer & 0.0080  & 0.0345  & 0.0748  & 0.2963   & 0.4113   & 0.8151 & 0.7058  & 6.77  & 7.46  & 9.96  & 0.81  & 0.0825   \\ 
        DML & 0.0167  & 0.0669  & 0.1296  & 0.3154   & 0.4221  & 0.8164   & 0.7069  &6.97 & 7.87  & 10.50  & 0.50  & 0.1427    \\
        CL-ILR & 0.0167  & 0.0686  & 0.1369  & 0.3223  & 0.4241    & 0.8179    & 0.7078  & 6.95  & 7.87  & 10.50  & 0.51  & 0.1355    \\ 
        ONE & 0.0120  & 0.0489  & 0.0995  & 0.3248  & 0.4082    & 0.8170   & 0.7072  & 6.95  & 7.76  & 10.42  & 0.66  & 0.1174    \\ 
        OKDDip & 0.0141  & 0.0581  & 0.1168  & 0.3097  & 0.4212    & \underline{0.8188}    & 0.7100  & 6.90  & 7.75  & 10.31  & 0.55  & 0.1376  \\  \midrule
        CDL-CI &  \underline{0.0191}  & \underline{0.0815}  & \underline{0.1679}  & 0.3139   & \textbf{0.4283}   & 0.8182   & 0.7074  & 6.89  & 7.83  & 10.45  & \underline{0.41}  & 0.1514    \\ 
        CDL-CS & 0.0186  & 0.0785  & 0.1561  & \textbf{0.3317}  & 0.4108   & \textbf{0.8198}  & \textbf{0.7177}  & \underline{7.07}  & \underline{7.99}  & \underline{10.70}  & 0.42  & \underline{0.1603}    \\ 
        CDL-IS & \textbf{0.0252}  & \textbf{0.1081}  & \textbf{0.2143}  & \underline{0.3184}   & \underline{0.4261}   & 0.8179  & \underline{0.7121}  & \textbf{7.12}  & \textbf{8.13}  & \textbf{10.80}  & \textbf{0.32}  & \textbf{0.1778}  \\  \bottomrule
        \toprule
       Transformer & 0.0031  & 0.0140  & 0.0302  & 0.3552  & 0.3062  & 0.7891  & \underline{0.7048}  & 6.71  & 7.64  & \underline{10.99}  & 1.31  & 0.0349  \\ 
        DML & 0.0044  & 0.0171  & 0.0344  & 0.3494   & 0.3248  & 0.7907  & 0.6801  & 6.41  & 7.11  & 10.16  & 1.58  & 0.0363   \\
        CL-ILR & 0.0044  & 0.0179  & 0.0368  & 0.3310   & 0.3151  & 0.7804  & 0.6648  & 6.48  & 7.26  & 10.32  & 1.49  & 0.0513  \\
        ONE & 0.0043  & 0.0175  & 0.0369  & 0.3510   & 0.3140  & 0.7922  & 0.6921  & 6.56  & 7.43  & 10.60  & 1.40  & 0.0410   \\
        OKDDip & 0.0035  & 0.0141  & 0.0300  & 0.3487  & 0.3244  & 0.7886  & 0.6743  & 6.49  & 7.19  & 10.30  & 1.55  & 0.0356   \\ \midrule
        CDL-CI & \textbf{0.0057}  & \textbf{0.0239}  & \textbf{0.0523}  & \underline{0.3474}   & \underline{0.3254}  & \textbf{0.7983}  & \textbf{0.7156}  & \underline{6.73}  & \textbf{7.71}  & \textbf{11.03}  & \textbf{1.18} & \textbf{0.0555}   \\ 
        CDL-CS & \underline{0.0050}  & 0.0197  & 0.0419  & \textbf{0.3552} & 0.3146  & \underline{0.7924}  & 0.6996  & 6.71  & 7.63  & 10.93  & \underline{1.29} & 0.0426   \\ 
        CDL-IS & \underline{0.0050}  & \underline{0.0211}  & \underline{0.0460}  & 0.3443  & \textbf{0.3258}  & 0.7893  & 0.6923  & \textbf{6.78}  & \underline{7.68}  & 10.95  & \underline{1.29} & \underline{0.0524}   \\ \bottomrule
    \end{tabular}
    \caption{Automatic evaluation results on DailyDialog (Up) and OpenSubtitles (Down). The best/second-best results are \textbf{bold}/\underline{underlined}. The branch number is 3. C refers to coherence, I for informativeness and S for specificity.}
    \label{tb:main_exp}
\end{table*}

\subsection{Implementation Details}
All approaches are based on the Transformer-based sequence-to-sequence model \citep{Transformer-Vaswani-2017}. Each branch is built on the lightweight model architecture (Small Transformer): the encoder and decoder contain only 2 layers, in which the self-attention module has 4 attention heads and 1024 feed-forward units. The size of hidden states is set to 256.
Dropout \citep{DBLP:journals/jmlr/SrivastavaHKSS14} is used for the self-attention module, the feed-forward layer, and the activation layer, and the rate of all three is set to 0.1.
The batch size is set to 64.
The selection ratio for attribute-specific subset is 70$\%$.
For the temperature coefficient $t$, we simply set it to 1. 
Beam search with a size of 5 is used for decoding.
We implement all approaches with Pytorch 1.11, and conduct all experiments on NVIDIA TITAN RTX.

\subsection{Comparison Methods}
We compare our proposed collaborative dialogue learning (CDL) framework with following established collaborative learning approaches:
\begin{itemize}
    \item DML \citep{DML} uses a pool of network-based students, where each student is an individual network and they asynchronously collaborate.
    \item CL-ILR \citep{CL-ILR} distills knowledge among multiple branches of a hierarchical network.
    \item ONE \citep{ONE} automatically generates gated ensemble logit from each branch as a soft target.
    \item OKDDip \citep{OKDDip} proposes a two-level distillation strategy with multiple auxiliary peers and a group leader, while utilizing an attention module to construct inter-branch diversity.
\end{itemize}

Following previous work, we set the branch number is 3 for all the comparison models. For the proposed framework, it contains one master branch and two auxiliary branches with different dialogue attributes, i.e.,  coherency (C), informativeness (I) and specificity (S).

\subsection{Automatic Evaluation}
\paragraph{Metrics}
We first used automatic metrics to evaluate our method:
\textbf{Dist-\{1,2,3\}} (distinct) \citep{MMI-LiJiwei-2016} is a widely used metric that reflects the lexical diversity of the generated responses by calculating the proportion of unique unigrams/bigrams/trigrams.
\textbf{BLEU} \citep{DBLP:conf/wmt/ChenC14} measures n-gram overlap between the generated and the ground-truth responses. 
\textbf{AVE} (Embedding Average) \citep{DBLP:conf/emnlp/LiuLSNCP16} evaluates the semantic relationship of generated responses and ground-truth responses.
\textbf{COH} (coherence) \citep{FilterCoherence-Xu-2018} measures the cosine similarity between pairs of input and response. 
\textbf{H-\{1,2,3\}} (word entropy) \citep{VHRED-Serban-2017} measures the unigrams/bigrams/trigrams' non-genericness of responses.
\textbf{KL} (KL divergence) \citep{FilterEntropy-Csaky-2019} measures the distribution distance between the generated and the ground-truth response sets to reflect how well a model can approximate the ground-truth distribution. Note that the lower KL is better.
\textbf{LF} (low-frequency token ratio) \citep{DBLP:conf/iclr/LiWCUS0Z20} further measures the diversity of responses by calculating the ratio of low-frequency words in the generated responses. The threshold of low frequency is set to 100.

\paragraph{Results}
Table~\ref{tb:main_exp} shows the results obtained at the lowest point of the validation loss. It illustrates that our framework outperforms all baselines by a significant margin on both datasets. Note that four collaborative learning baselines perform better than vanilla Transformer model, which proves that the group-base distillation can improve model performance greatly. On this basis, the proposed approach can further enhance the performance by introducing dialogue attributes learning and dual knowledge distillation. And the improvement of different dialogue attributes can be reflected by the corresponding metrics.
\subsection{Human Evaluation}
\begin{table}[th]
    \centering
    \begin{tabular}{l c c c c}
    \toprule
        vs. Models & Win & Tie & Loss & Kappa  \\  \midrule
        Transformer & 0.82 & 0.15 & 0.03 & 0.5487\\
        DML  & 0.47 & 0.42 & 0.11 & 0.6651 \\
        CL-ILR  & 0.43 & 0.53 & 0.05 &  0.5393\\
        ONE  & 0.50 & 0.39 & 0.11 & 0.6177\\
        OKDDip & 0.45 & 0.43 & 0.11 & 0.5743\\ 
        \bottomrule
    \end{tabular}
    \caption{Human evaluations results on DailyDialog. Our framework has a higher win rate than baselines.}
    \label{tb:human}
\end{table}

To further verify the effectiveness of our method in comparison to previous collaborative learning methods, we also conduct human evaluations apart from automatic evaluations.
We randomly select 50 samples from the test set of DailyDialog, and three well-educated annotators are invited to judge which of the overall response quality generated by CDL and baselines is better (i.e., win, tie or loss) in terms of coherence, informativeness and fluency.

Table~\ref{tb:human} summarizes the human evaluation results. In our experience, we have noticed that a dialogue model trained using our proposed learning framework is more capable of producing responses that are human-preferred.
We use Fleiss's kappa \citep{fleisskappa/measuring} to measure the inter-annotator agreement, which indicates that the annotators came to a fair agreement in the judgment.

\subsection{Analysis}
In order to better understand the effectiveness of the collaborative dialogue learning, we carry out extensive analysis of DailyDialog.

\paragraph{Ablation study} 
\begin{table}[th]
    \centering
    \begin{tabular}{l c c c c c c c c}
    \toprule
        Models & Dist-1 & Dist-2 & LF  & KL  & H-1 \\  \midrule
        w/o Attributes & 0.0177 & 0.0756 & 0.1367 & 0.38  & 6.94  \\
        w/o OP & 0.0203 & 0.0869 & 0.1629 & 0.37 & 6.99 \\
        w/o $\mathcal{L}_{ND_{hidden}}$  & 0.0222 & 0.0931 & 0.1621 & 0.41 & 7.03 \\
        w/o $\mathcal{L}_{neg}$  & 0.0215 & 0.0890 & 0.1776 & 0.43 & 7.05 \\
 \midrule
        Full Version   & 0.0252 & 0.1081 & 0.1778 & 0.32 & 7.12 \\
        \bottomrule
    \end{tabular}
    \caption{Ablation study results of the proposed collaborative dialogue learning framework.}
    \label{tb:ablation}
\end{table}

We study the effects of different parts of proposed framework by ablating the dialogue attribute learning (w/o attributes), the orthogonal projection (w/o OP), the hidden state distillation
(w/o $\mathcal{L}_{ND_{hidden}}$), and the whole negative distillation (w/o $\mathcal{L}_{Neg}$). 
The results in Table~\ref{tb:ablation} show that all proposed techniques are useful for improving the response quality. 
The significant decline in w/o attributes indicates that the knowledge of specific dialogue property is very important for CDL. 
w/o $\mathcal{L}_{Neg}$ is better than w/o OP, indicating that orthogonal projection is a key technique to capture biased features without harming common knowledge.

\paragraph{Comparison with traditional KD}

\begin{table}[th]
    \centering
    \begin{tabular}{l c c c c c c c c}
    \toprule
        Models & Dist-1 & Dist-2 & LF  & KL  & H-1 \\  \midrule
        Small & 0.0080 & 0.0345 & 0.0825 & 0.81 & 6.77 \\
        Base & 0.0101  & 0.0471  & 0.1084  & 0.56  & 6.83  \\
        KD & 0.0124  & 0.0564  & 0.1336  & 0.47  & 6.94  \\
        CDL-IS & 0.0252 & 0.1081 & 0.1778 & 0.32 & 7.12 \\
        \bottomrule
    \end{tabular}
    \caption{Comparison results with traditional knowledge distillation.}
    \label{tb:kd}
\end{table}

Traditional knowledge distillation is an efficient method to obtain a small but effective model. The results from Table~\ref{tb:kd} show that CDL outperform KD (Teacher is Base Transformer) and Small with the same inference cost and the relative heavy Base model without a well-trained teacher.

\paragraph{Branch Number Study}
\begin{table}[th]
    \centering
    \begin{tabular}{l c c c c c c c c}
    \toprule
        Models & Dist-1 & Dist-2 & LF  & KL  & H-1 \\  \midrule
        DML   & 0.0154 & 0.0644 & 0.1368  & 0.51  & 6.95  \\
        CL-ILR  & 0.0154 & 0.0625 & 0.1232  & 0.52  & 6.90  \\
        ONE  & 0.0104 & 0.0432 & 0.1022  & 0.69  & 6.88  \\
        OKDDip  & 0.0132 & 0.0576 & 0.1348  & 0.48  & 6.96  \\
        CDL-C  & 0.0195 & 0.0833 & 0.1492 & 0.39 & 6.97  \\
        CDL-S & 0.0182 & 0.0730 & 0.1515 & 0.54  & \textbf{7.03}  \\
        CDL-I  & \textbf{0.0207} & \textbf{0.0889} & \textbf{0.1834} & \textbf{0.37}  & 6.93  \\
        \bottomrule
        \toprule
        Models & Dist-1 & Dist-2 & LF  & KL  & H-1 \\  \midrule
        DML   & 0.0198 & 0.0779 & 0.1457  & 0.44  & 6.93  \\
        CL-ILR  & 0.0159 & 0.0637 & 0.1362  & 0.56  & 6.97  \\
        ONE  & 0.0128 & 0.0523 & 0.1194  & 0.61  & 6.92  \\
        OKDDip  & 0.0167 & 0.0655 & 0.1274  & 0.56  & 6.99  \\
        CDL-CSI  & \textbf{0.0211} & \textbf{0.0869} & \textbf{0.1672} & \textbf{0.41}   & \textbf{6.99} \\
        \bottomrule
    \end{tabular}
    \caption{Evaluation Results with branch number 2 (Up) and 4 (Down).}
    \label{tb:number}
\end{table}

We explore the performance of proposed CDL in other number of branches. The results from Table~\ref{tb:number} shows that, regardless of the number of branches, performance of CDL is better than baselines. The inferior performance with branch number 4 (compared with Table~\ref{tb:main_exp}) is that the influence of positive distillation from master branch is much lower than with branch number 3, given that the more auxiliary branches the more negative distillation will be conducted. The promising way to solve this problem is to increase the weight of positive KD loss and decrease negative KD’s, and we leave it as future work.

\paragraph{Model Diversity Analysis}

\begin{table}[th]
    \centering
    \begin{tabular}{l c c c c c}
    \toprule
         & DML & CL-ILR & ONE  & OKDDip & CDL  \\  \midrule
        L2  & 0.163 & 0.141 & 0.198 & 0.165 & 0.308 \\
        \bottomrule
    \end{tabular}
    \caption{Branches diversity for CDL and other collaborative baselines.}
    \label{tb:distance}
\end{table}

we show the diversity that our method brings in a more intuitive way. We use the Euclidean distance (L2) between branches as a quantitative criterion for diversity, as performed in OKDDip \citep{OKDDip}. Table~\ref{tb:distance} shows the averaged L2 distances when each model reaches to convergence. Our mutual diversity is significantly greater than other methods, which suggests that the homogenization problem between branches have greatly relieved through multi-attribute learning and negative distillation.

\paragraph{Case Study}

\begin{table}[ht]
\centering

\begin{tabular}{@{}p{\linewidth}@{}}
\toprule
\textbf{Input:} Here are all kinds of jades. Choose whatever you like, please. \\
\midrule
\textbf{Transformer:} Ok. \\
\textbf{DML:} Thank you. \\
\textbf{CL-ILR:} Thank you very much. \\
\textbf{ONE:} Thank you. I have a good idea. \\
\textbf{OKDDip:} Thank you very much.  \\
\textbf{CDL:} Thank you, sir. I 'll be glad to do that.  \\
\midrule
\midrule
\textbf{Input:} What do you recommend? We'd like some of your local specialities. \\
\midrule
\textbf{Transformer:} That's good. \\
\textbf{DML:} That's a good idea. What do you want to do? \\
\textbf{CL-ILR:} I don't know what you mean. \\
\textbf{ONE:} Well, I think I'd like to have a cup of coffee. \\
\textbf{OKDDip:} Well, I've heard about it, but I don't think it's a good idea.  \\
\textbf{CDL:} That's great. I'd like to have some cheese.  \\
\bottomrule
\end{tabular}
\caption{Examples of generated responses.}
\label{tb:case study}
\end{table}

Table \ref{tb:case study} presents some responses generated by the proposed framework and baselines. Transformer prefers generic and meaningless responses. Other baselines lack of concerning one perspective of dialogue. In contrast, our CDL comprehensively consider the multiple perspectives, thus resulting in diverse and coherent responses. The results demonstrate the effectiveness of CDL.
\section{Related Work}
\subsection{Dialogue Models}
There are three major categories of previous work on enhancing the quality of responses. 
The first redesigns the model structure to facilitate the modeling of the dialogue pairs \citep{DBLP:conf/aaai/SerbanKTTZBC17,CMHAM-TaoChongyang-2018,SpaceFusion-GaoXiang-2019}. 
The second further proposes advanced objectives aligning with the goals of the conversation more effectively, such as MMI \citep{MMI-LiJiwei-2016}, CVAE \citep{VHRED-Serban-2017,kgCVAE-ZhaoTiancheng-2017,DialogWAE-GuXiaodong-2019,DBLP:conf/acl/SunFLLL20}, RL \citep{RLdialoguesys-LiJiwei-2016,RL-Seq2seqCo-Zhang2018,RL-P2BOT-Liu2020}, and GAN \citep{GAN-GANAEL-Xu2017,DPGAN-XuJingjing-2018,PosteriorGan-FengShaoxiong2020}. 
The third tries to endow the responses with topic \citep{TopicAware-XingChen-2017,DBLP:conf/emnlp/FengRCSLS20}, emotion \cite{Emotional-Zhouhao-2018,DBLP:conf/acl/RashkinSLB19}, and persona \citep{Persona-QianQiao-2017,PersonaChat-facebook-2018,DBLP:conf/acl/SongWZLL20}.
Recently, data filtering has been introduced for dialogue learning, which discards samples regarded as low-quality by a scoring method to reflect corresponding dialogue attributes. 
\citet{FilterEntropy-Csaky-2019} proposes an entropy-based scoring method to remove generic utterances from the training data. \citet{FilterSpecificity-See-2019} designs a scoring method to measure the specificity of samples. \citet{FilterConsistency-Akama-2020} combines the cosine distance and the keyword co-occurrence of the dialogue pairs to evaluate the coherence. \citet{DBLP:conf/cikm/ShenZSCZZ21} presents a fusing approach to data filtering and \citet{DBLP:journals/corr/abs-2205-11206} utilizes the scoring methods to enhance rather than filter data.

\subsection{Knowledge Distillation}
In recent years, knowledge distillation \citep{DBLP:journals/corr/HintonVD15, DBLP:journals/corr/FreitagAS17} has been widely adopted by researchers to accelerate and compress models \citep{DBLP:conf/emnlp/JiaoYSJCL0L20, DBLP:journals/corr/abs-1910-01108}. As these predicted distributions contain ranking information on similarities among categories, it treats the predictions as knowledge learned by the teacher network. As a result, it enforces similar predictions on the student network in order to transfer this knowledge. By providing more knowledge to the student network from different sources, the work follows this idea. To supervise the student network, FitNets \citep{DBLP:journals/corr/RomeroBKCGB14} uses both predictions and intermediate representations learned by the teacher network. \citet{DBLP:conf/emnlp/KimR16} propose using sequence-level knowledge generated from the generated sequences to guide student network training in the Seq2Seq model. Furthermore, self-knowledge distribution \citep{DBLP:conf/ranlp/HahnC19} demonstrates that students are able to improve performance by using their own knowledge. When it comes to dialogue generation, \citet{multi-view} guide the dialogue model towards better generalization by introducing bidirectional distillation and \citet{negative-distill} propose negative distillation to enhance the diversity of responses.
Rather than pre-training a large teacher, we use collaborative learning and distill knowledge from a group of branches. 

\subsection{Collaborative Learning}
The concept of collaborative learning \citep{Online-distill,ONE,CGL,CL-ILR,OKDDip} is more lightweight in terms of the stages of learning compared to that of conventional knowledge distillation. By doing so, it facilitates finding a robust local minimum for each student, resulting in greater generalization performance. Student networks are currently being implemented in two mainstream settings. One is network-based \citep{DML}, in which students form independent networks and parameter capacity increases linearly with the number of students; the other is branch-based (CL-ILR \citep{CL-ILR} and ONE \citep{ONE}), where the bottom layers of students are shared. \citet{CGL} enable more flexible representation sharing with random routing mechanism, where layers at any level can be shared by different involved students. However, previous work focused only on classification, resulting in the same training objective for all branches of the framework with independent identical distributions (i.i.d.). As a result, different branches will tend to converge to similar feature representations \citep{Convergent,ONE,OKDDip}. Different from them, we propose attrbute-related branch learning strategy and dual knowledge distillation to solve the homogenization problem.
\section{Conclusion}
We present a novel collaborative dialogue learning paradigm to improve the quality of generated responses in terms of three major dialogue attributes. 
CDL replaces traditional knowledge distillation with collaborative group-based distillation for lightweight knowledge interaction, and the attribute-aware knowledge is captured and transferred through auxiliary branches.
Dual group-based knowledge distillation is proposed for better guiding auxiliary branches to learn attribute-specific knowledge.
Besides, we further boost the performance of negative distillation by utilizing orthogonal projection to avoid harming the common knowledge.
Extensive experiments validate the superiority of our proposed method over prior collaborative learning work.

\section{Acknowledgments}
This research was supported by the Beijing Natural Science Foundation (No.4222037, L181010) and the BIT Research and Innovation Promoting Project (Grant No.2022YCXY021).

\bibliography{aaai23}

\begin{thebibliography}{58}
\providecommand{\natexlab}[1]{#1}

\bibitem[{Akama et~al.(2020)Akama, Yokoi, Suzuki, and
  Inui}]{FilterConsistency-Akama-2020}
Akama, R.; Yokoi, S.; Suzuki, J.; and Inui, K. 2020.
\newblock Filtering Noisy Dialogue Corpora by Connectivity and Content
  Relatedness.
\newblock In \emph{{EMNLP}}, 941--958.

\bibitem[{Anil et~al.(2018)Anil, Pereyra, Passos, Orm{\'{a}}ndi, Dahl, and
  Hinton}]{Online-distill}
Anil, R.; Pereyra, G.; Passos, A.; Orm{\'{a}}ndi, R.; Dahl, G.~E.; and Hinton,
  G.~E. 2018.
\newblock Large scale distributed neural network training through online
  distillation.
\newblock In \emph{{ICLR}}.

\bibitem[{Bouma(2009)}]{nPMI-Bouma-2009}
Bouma, G. 2009.
\newblock Normalized (Pointwise) Mutual Information in Collocation Extraction.
\newblock In \emph{(GSCL)}, 31–40.

\bibitem[{Chen and Cherry(2014)}]{DBLP:conf/wmt/ChenC14}
Chen, B.; and Cherry, C. 2014.
\newblock A Systematic Comparison of Smoothing Techniques for Sentence-Level
  {BLEU}.
\newblock In \emph{Ninth Workshop on Statistical Machine Translation},
  362--367.

\bibitem[{Chen et~al.(2020)Chen, Mei, Wang, Feng, and Chen}]{OKDDip}
Chen, D.; Mei, J.; Wang, C.; Feng, Y.; and Chen, C. 2020.
\newblock Online Knowledge Distillation with Diverse Peers.
\newblock In \emph{{AAAI}}, 3430--3437.

\bibitem[{Csaky, Purgai, and Recski(2019)}]{FilterEntropy-Csaky-2019}
Csaky, R.; Purgai, P.; and Recski, G. 2019.
\newblock Improving Neural Conversational Models with Entropy-Based Data
  Filtering.
\newblock In \emph{{ACL} {(1)}}, 5650--5669.

\bibitem[{Feng et~al.(2020{\natexlab{a}})Feng, Chen, Li, and
  Yin}]{PosteriorGan-FengShaoxiong2020}
Feng, S.; Chen, H.; Li, K.; and Yin, D. 2020{\natexlab{a}}.
\newblock Posterior-GAN: Towards Informative and Coherent Response Generation
  with Posterior Generative Adversarial Network.
\newblock In \emph{{AAAI}}, 7708--7715.

\bibitem[{Feng et~al.(2021{\natexlab{a}})Feng, Chen, Ren, Ding, Li, and
  Sun}]{CGL}
Feng, S.; Chen, H.; Ren, X.; Ding, Z.; Li, K.; and Sun, X. 2021{\natexlab{a}}.
\newblock Collaborative Group Learning.
\newblock In \emph{{AAAI}}, 7431--7438.

\bibitem[{Feng et~al.(2020{\natexlab{b}})Feng, Ren, Chen, Sun, Li, and
  Sun}]{DBLP:conf/emnlp/FengRCSLS20}
Feng, S.; Ren, X.; Chen, H.; Sun, B.; Li, K.; and Sun, X. 2020{\natexlab{b}}.
\newblock Regularizing Dialogue Generation by Imitating Implicit Scenarios.
\newblock In \emph{{EMNLP}}, 6592--6604.

\bibitem[{Feng et~al.(2021{\natexlab{b}})Feng, Ren, Li, and Sun}]{multi-view}
Feng, S.; Ren, X.; Li, K.; and Sun, X. 2021{\natexlab{b}}.
\newblock Multi-View Feature Representation for Dialogue Generation with
  Bidirectional Distillation.
\newblock In \emph{{AAAI}}, 12812--12820.

\bibitem[{Fleiss(1971)}]{fleisskappa/measuring}
Fleiss, J.~L. 1971.
\newblock Measuring nominal scale agreement among many raters.
\newblock \emph{Psychological bulletin}, 76(5): 378.

\bibitem[{Freitag, Al{-}Onaizan, and
  Sankaran(2017)}]{DBLP:journals/corr/FreitagAS17}
Freitag, M.; Al{-}Onaizan, Y.; and Sankaran, B. 2017.
\newblock Ensemble Distillation for Neural Machine Translation.
\newblock \emph{CoRR}.

\bibitem[{Gao et~al.(2019)Gao, Lee, Zhang, Brockett, Galley, Gao, and
  Dolan}]{SpaceFusion-GaoXiang-2019}
Gao, X.; Lee, S.; Zhang, Y.; Brockett, C.; Galley, M.; Gao, J.; and Dolan, B.
  2019.
\newblock Jointly Optimizing Diversity and Relevance in Neural Response
  Generation.
\newblock In \emph{{NAACL-HLT} {(1)}}, 1229--1238.

\bibitem[{Gu et~al.(2019)Gu, Cho, Ha, and Kim}]{DialogWAE-GuXiaodong-2019}
Gu, X.; Cho, K.; Ha, J.; and Kim, S. 2019.
\newblock DialogWAE: Multimodal Response Generation with Conditional
  Wasserstein Auto-Encoder.
\newblock In \emph{{ICLR} (Poster)}.

\bibitem[{Hahn and Choi(2019)}]{DBLP:conf/ranlp/HahnC19}
Hahn, S.; and Choi, H. 2019.
\newblock Self-Knowledge Distillation in Natural Language Processing.
\newblock In \emph{{RANLP}}, 423--430.

\bibitem[{Hinton, Vinyals, and Dean(2015)}]{DBLP:journals/corr/HintonVD15}
Hinton, G.~E.; Vinyals, O.; and Dean, J. 2015.
\newblock Distilling the Knowledge in a Neural Network.
\newblock \emph{CoRR}, abs/1503.02531.

\bibitem[{Jiao et~al.(2020)Jiao, Yin, Shang, Jiang, Chen, Li, Wang, and
  Liu}]{DBLP:conf/emnlp/JiaoYSJCL0L20}
Jiao, X.; Yin, Y.; Shang, L.; Jiang, X.; Chen, X.; Li, L.; Wang, F.; and Liu,
  Q. 2020.
\newblock TinyBERT: Distilling {BERT} for Natural Language Understanding.
\newblock In \emph{Findings of {EMNLP}}, 4163--4174.

\bibitem[{Kim and Rush(2016)}]{DBLP:conf/emnlp/KimR16}
Kim, Y.; and Rush, A.~M. 2016.
\newblock Sequence-Level Knowledge Distillation.
\newblock In \emph{{EMNLP}}, 1317--1327.

\bibitem[{Lan, Zhu, and Gong(2018)}]{ONE}
Lan, X.; Zhu, X.; and Gong, S. 2018.
\newblock Knowledge Distillation by On-the-Fly Native Ensemble.
\newblock In \emph{{NeurIPS}}, 7528--7538.

\bibitem[{Li et~al.(2016{\natexlab{a}})Li, Galley, Brockett, Gao, and
  Dolan}]{MMI-LiJiwei-2016}
Li, J.; Galley, M.; Brockett, C.; Gao, J.; and Dolan, B. 2016{\natexlab{a}}.
\newblock A Diversity-Promoting Objective Function for Neural Conversation
  Models.
\newblock In \emph{{HLT-NAACL}}, 110--119.

\bibitem[{Li et~al.(2016{\natexlab{b}})Li, Monroe, Ritter, Jurafsky, Galley,
  and Gao}]{RLdialoguesys-LiJiwei-2016}
Li, J.; Monroe, W.; Ritter, A.; Jurafsky, D.; Galley, M.; and Gao, J.
  2016{\natexlab{b}}.
\newblock Deep Reinforcement Learning for Dialogue Generation.
\newblock In \emph{{EMNLP}}, 1192--1202.

\bibitem[{Li et~al.(2022{\natexlab{a}})Li, Feng, Sun, and
  Li}]{negative-distill}
Li, Y.; Feng, S.; Sun, B.; and Li, K. 2022{\natexlab{a}}.
\newblock Diversifying Neural Dialogue Generation via Negative Distillation.
\newblock In \emph{{NAACL}}, 407--418.

\bibitem[{Li et~al.(2017)Li, Su, Shen, Li, Cao, and Niu}]{dailydialog2017}
Li, Y.; Su, H.; Shen, X.; Li, W.; Cao, Z.; and Niu, S. 2017.
\newblock DailyDialog: {A} Manually Labelled Multi-turn Dialogue Dataset.
\newblock In \emph{{IJCNLP(1)}}, 986--995.

\bibitem[{Li et~al.(2022{\natexlab{b}})Li, Sun, Feng, and
  Li}]{DBLP:journals/corr/abs-2205-11206}
Li, Y.; Sun, B.; Feng, S.; and Li, K. 2022{\natexlab{b}}.
\newblock Stop Filtering: Multi-View Attribute-Enhanced Dialogue Learning.
\newblock \emph{CoRR}, abs/2205.11206.

\bibitem[{Li et~al.(2016{\natexlab{c}})Li, Yosinski, Clune, Lipson, and
  Hopcroft}]{Convergent}
Li, Y.; Yosinski, J.; Clune, J.; Lipson, H.; and Hopcroft, J.~E.
  2016{\natexlab{c}}.
\newblock Convergent Learning: Do different neural networks learn the same
  representations?
\newblock In \emph{{ICLR}}.

\bibitem[{Li et~al.(2020)Li, Wang, Chen, Utiyama, Sumita, Zhang, and
  Zhao}]{DBLP:conf/iclr/LiWCUS0Z20}
Li, Z.; Wang, R.; Chen, K.; Utiyama, M.; Sumita, E.; Zhang, Z.; and Zhao, H.
  2020.
\newblock Data-dependent Gaussian Prior Objective for Language Generation.
\newblock In \emph{{ICLR}}.

\bibitem[{Liu et~al.(2016)Liu, Lowe, Serban, Noseworthy, Charlin, and
  Pineau}]{DBLP:conf/emnlp/LiuLSNCP16}
Liu, C.; Lowe, R.; Serban, I.; Noseworthy, M.; Charlin, L.; and Pineau, J.
  2016.
\newblock How {NOT} To Evaluate Your Dialogue System: An Empirical Study of
  Unsupervised Evaluation Metrics for Dialogue Response Generation.
\newblock In Su, J.; Carreras, X.; and Duh, K., eds., \emph{]{EMNLP}},
  2122--2132.

\bibitem[{Liu et~al.(2020)Liu, Chen, Chen, Lou, Chen, Zhou, and
  Zhang}]{RL-P2BOT-Liu2020}
Liu, Q.; Chen, Y.; Chen, B.; Lou, J.; Chen, Z.; Zhou, B.; and Zhang, D. 2020.
\newblock You Impress Me: Dialogue Generation via Mutual Persona Perception.
\newblock In \emph{{ACL}}, 1417--1427.

\bibitem[{Qian et~al.(2017)Qian, Huang, Zhao, Xu, and
  Zhu}]{Persona-QianQiao-2017}
Qian, Q.; Huang, M.; Zhao, H.; Xu, J.; and Zhu, X. 2017.
\newblock Assigning personality/identity to a chatting machine for coherent
  conversation generation.
\newblock \emph{CoRR}, abs/1706.02861.

\bibitem[{Rashkin et~al.(2019)Rashkin, Smith, Li, and
  Boureau}]{DBLP:conf/acl/RashkinSLB19}
Rashkin, H.; Smith, E.~M.; Li, M.; and Boureau, Y. 2019.
\newblock Towards Empathetic Open-domain Conversation Models: {A} New Benchmark
  and Dataset.
\newblock In \emph{{ACL}}, 5370--5381.

\bibitem[{Romero et~al.(2015)Romero, Ballas, Kahou, Chassang, Gatta, and
  Bengio}]{DBLP:journals/corr/RomeroBKCGB14}
Romero, A.; Ballas, N.; Kahou, S.~E.; Chassang, A.; Gatta, C.; and Bengio, Y.
  2015.
\newblock FitNets: Hints for Thin Deep Nets.
\newblock In Bengio, Y.; and LeCun, Y., eds., \emph{{ICLR}}.

\bibitem[{Sanh et~al.(2019)Sanh, Debut, Chaumond, and
  Wolf}]{DBLP:journals/corr/abs-1910-01108}
Sanh, V.; Debut, L.; Chaumond, J.; and Wolf, T. 2019.
\newblock DistilBERT, a distilled version of {BERT:} smaller, faster, cheaper
  and lighter.
\newblock \emph{CoRR}, abs/1910.01108.

\bibitem[{See et~al.(2019)See, Roller, Kiela, and
  Weston}]{FilterSpecificity-See-2019}
See, A.; Roller, S.; Kiela, D.; and Weston, J. 2019.
\newblock What makes a good conversation? How controllable attributes affect
  human judgments.
\newblock In \emph{{NAACL-HLT}}, 1702--1723.

\bibitem[{Serban et~al.(2017{\natexlab{a}})Serban, Klinger, Tesauro,
  Talamadupula, Zhou, Bengio, and Courville}]{DBLP:conf/aaai/SerbanKTTZBC17}
Serban, I.~V.; Klinger, T.; Tesauro, G.; Talamadupula, K.; Zhou, B.; Bengio,
  Y.; and Courville, A.~C. 2017{\natexlab{a}}.
\newblock Multiresolution Recurrent Neural Networks: An Application to Dialogue
  Response Generation.
\newblock In \emph{{AAAI}}, 3288--3294. {AAAI} Press.

\bibitem[{Serban et~al.(2017{\natexlab{b}})Serban, Sordoni, Lowe, Charlin,
  Pineau, Courville, and Bengio}]{VHRED-Serban-2017}
Serban, I.~V.; Sordoni, A.; Lowe, R.; Charlin, L.; Pineau, J.; Courville,
  A.~C.; and Bengio, Y. 2017{\natexlab{b}}.
\newblock A Hierarchical Latent Variable Encoder-Decoder Model for Generating
  Dialogues.
\newblock In \emph{{AAAI}}, 3295--3301.

\bibitem[{Shang, Lu, and Li(2015)}]{Seq2Seq-ShangLifeng-2015}
Shang, L.; Lu, Z.; and Li, H. 2015.
\newblock Neural Responding Machine for Short-Text Conversation.
\newblock In \emph{{ACL} {(1)}}, 1577--1586.

\bibitem[{Shen et~al.(2021)Shen, Zhan, Shen, Chen, Zhao, and
  Zhu}]{DBLP:conf/cikm/ShenZSCZZ21}
Shen, L.; Zhan, H.; Shen, X.; Chen, H.; Zhao, X.; and Zhu, X. 2021.
\newblock Identifying Untrustworthy Samples: Data Filtering for Open-domain
  Dialogues with Bayesian Optimization.
\newblock In \emph{{CIKM}}, 1598--1608.

\bibitem[{Song and Chai(2018)}]{CL-ILR}
Song, G.; and Chai, W. 2018.
\newblock Collaborative Learning for Deep Neural Networks.
\newblock In \emph{{NeurIPS}}, 1837--1846.

\bibitem[{Song et~al.(2020)Song, Wang, Zhang, Liu, and
  Liu}]{DBLP:conf/acl/SongWZLL20}
Song, H.; Wang, Y.; Zhang, W.; Liu, X.; and Liu, T. 2020.
\newblock Generate, Delete and Rewrite: {A} Three-Stage Framework for Improving
  Persona Consistency of Dialogue Generation.
\newblock In \emph{{ACL}}, 5821--5831.

\bibitem[{Sordoni et~al.(2015)Sordoni, Galley, Auli, Brockett, Ji, Mitchell,
  Nie, Gao, and Dolan}]{Seq2Seq-Sordoni-2015}
Sordoni, A.; Galley, M.; Auli, M.; Brockett, C.; Ji, Y.; Mitchell, M.; Nie, J.;
  Gao, J.; and Dolan, B. 2015.
\newblock A Neural Network Approach to Context-Sensitive Generation of
  Conversational Responses.
\newblock In \emph{{HLT-NAACL}}, 196--205.

\bibitem[{Srivastava et~al.(2014)Srivastava, Hinton, Krizhevsky, Sutskever, and
  Salakhutdinov}]{DBLP:journals/jmlr/SrivastavaHKSS14}
Srivastava, N.; Hinton, G.~E.; Krizhevsky, A.; Sutskever, I.; and
  Salakhutdinov, R. 2014.
\newblock Dropout: a simple way to prevent neural networks from overfitting.
\newblock \emph{J. Mach. Learn. Res.}, 15(1): 1929--1958.

\bibitem[{Sun et~al.(2021)Sun, Feng, Li, Liu, and Li}]{DBLP:conf/acl/SunFLLL20}
Sun, B.; Feng, S.; Li, Y.; Liu, J.; and Li, K. 2021.
\newblock Generating Relevant and Coherent Dialogue Responses using
  Self-Separated Conditional Variational AutoEncoders.
\newblock In \emph{{ACL/IJCNLP}}, 5624--5637.

\bibitem[{Sun et~al.(2019)Sun, Cheng, Gan, and Liu}]{DBLP:conf/emnlp/SunCGL19}
Sun, S.; Cheng, Y.; Gan, Z.; and Liu, J. 2019.
\newblock Patient Knowledge Distillation for {BERT} Model Compression.
\newblock In \emph{{EMNLP-IJCNLP}}, 4322--4331.

\bibitem[{Tao et~al.(2018)Tao, Gao, Shang, Wu, Zhao, and
  Yan}]{CMHAM-TaoChongyang-2018}
Tao, C.; Gao, S.; Shang, M.; Wu, W.; Zhao, D.; and Yan, R. 2018.
\newblock Get The Point of My Utterance! Learning Towards Effective Responses
  with Multi-Head Attention Mechanism.
\newblock In \emph{{IJCAI}}, 4418--4424.

\bibitem[{Tiedemann(2009)}]{opensubtitles2009}
Tiedemann, J. 2009.
\newblock \emph{News from OPUS—A Collection of Multilingual Parallel Corpora
  with Tools and Interfaces}.

\bibitem[{Vaswani et~al.(2017)Vaswani, Shazeer, Parmar, Uszkoreit, Jones,
  Gomez, Kaiser, and Polosukhin}]{Transformer-Vaswani-2017}
Vaswani, A.; Shazeer, N.; Parmar, N.; Uszkoreit, J.; Jones, L.; Gomez, A.~N.;
  Kaiser, L.; and Polosukhin, I. 2017.
\newblock Attention is All you Need.
\newblock In \emph{{NIPS}}, 5998--6008.

\bibitem[{Vinyals and Le(2015)}]{NoisyData-Vinyals-2015}
Vinyals, O.; and Le, Q.~V. 2015.
\newblock A Neural Conversational Model.
\newblock In \emph{{ICML} Deep Learning Workshop}.

\bibitem[{Wang et~al.(2019)Wang, He, Lipton, and
  Xing}]{DBLP:conf/iclr/WangHLX19}
Wang, H.; He, Z.; Lipton, Z.~C.; and Xing, E.~P. 2019.
\newblock Learning Robust Representations by Projecting Superficial Statistics
  Out.
\newblock In \emph{{ICLR}}.

\bibitem[{Wu and Gong(2021)}]{PCL}
Wu, G.; and Gong, S. 2021.
\newblock Peer Collaborative Learning for Online Knowledge Distillation.
\newblock In \emph{{AAAI}}, 10302--10310.

\bibitem[{Xing et~al.(2017)Xing, Wu, Wu, Liu, Huang, Zhou, and
  Ma}]{TopicAware-XingChen-2017}
Xing, C.; Wu, W.; Wu, Y.; Liu, J.; Huang, Y.; Zhou, M.; and Ma, W. 2017.
\newblock Topic Aware Neural Response Generation.
\newblock In \emph{{AAAI}}, 3351--3357.

\bibitem[{Xu et~al.(2018{\natexlab{a}})Xu, Ren, Lin, and
  Sun}]{DPGAN-XuJingjing-2018}
Xu, J.; Ren, X.; Lin, J.; and Sun, X. 2018{\natexlab{a}}.
\newblock Diversity-Promoting {GAN:} {A} Cross-Entropy Based Generative
  Adversarial Network for Diversified Text Generation.
\newblock In \emph{{EMNLP}}, 3940--3949.

\bibitem[{Xu et~al.(2018{\natexlab{b}})Xu, Dusek, Konstas, and
  Rieser}]{FilterCoherence-Xu-2018}
Xu, X.; Dusek, O.; Konstas, I.; and Rieser, V. 2018{\natexlab{b}}.
\newblock Better Conversations by Modeling, Filtering, and Optimizing for
  Coherence and Diversity.
\newblock In \emph{{EMNLP}}, 3981--3991.

\bibitem[{Xu et~al.(2017)Xu, Liu, Wang, Sun, Wang, Wang, and
  Qi}]{GAN-GANAEL-Xu2017}
Xu, Z.; Liu, B.; Wang, B.; Sun, C.; Wang, X.; Wang, Z.; and Qi, C. 2017.
\newblock Neural Response Generation via {GAN} with an Approximate Embedding
  Layer.
\newblock In \emph{{EMNLP}}, 617--626.

\bibitem[{Zhang et~al.(2018{\natexlab{a}})Zhang, Lan, Guo, Xu, and
  Cheng}]{RL-Seq2seqCo-Zhang2018}
Zhang, H.; Lan, Y.; Guo, J.; Xu, J.; and Cheng, X. 2018{\natexlab{a}}.
\newblock Reinforcing Coherence for Sequence to Sequence Model in Dialogue
  Generation.
\newblock In \emph{{IJCAI}}, 4567--4573.

\bibitem[{Zhang et~al.(2018{\natexlab{b}})Zhang, Dinan, Urbanek, Szlam, Kiela,
  and Weston}]{PersonaChat-facebook-2018}
Zhang, S.; Dinan, E.; Urbanek, J.; Szlam, A.; Kiela, D.; and Weston, J.
  2018{\natexlab{b}}.
\newblock Personalizing Dialogue Agents: {I} have a dog, do you have pets too?
\newblock In \emph{{ACL}}, 2204--2213.

\bibitem[{Zhang et~al.(2018{\natexlab{c}})Zhang, Xiang, Hospedales, and
  Lu}]{DML}
Zhang, Y.; Xiang, T.; Hospedales, T.~M.; and Lu, H. 2018{\natexlab{c}}.
\newblock Deep Mutual Learning.
\newblock In \emph{{CVPR}}, 4320--4328.

\bibitem[{Zhao, Zhao, and Esk{\'{e}}nazi(2017)}]{kgCVAE-ZhaoTiancheng-2017}
Zhao, T.; Zhao, R.; and Esk{\'{e}}nazi, M. 2017.
\newblock Learning Discourse-level Diversity for Neural Dialog Models using
  Conditional Variational Autoencoders.
\newblock In \emph{{ACL} {(1)}}, 654--664.

\bibitem[{Zhou et~al.(2018)Zhou, Huang, Zhang, Zhu, and
  Liu}]{Emotional-Zhouhao-2018}
Zhou, H.; Huang, M.; Zhang, T.; Zhu, X.; and Liu, B. 2018.
\newblock Emotional Chatting Machine: Emotional Conversation Generation with
  Internal and External Memory.
\newblock In \emph{{AAAI-18}}, 730--739.

\end{thebibliography}

\end{document}